\documentclass{article}
\usepackage{spconf,amsmath,amssymb,graphicx}
\usepackage{booktabs}
\newcommand{\method}{\textsc{EgoSIS}}
\newcommand{\fvet}{\textsc{FVET}}
\newcommand{\retem}{\textsc{ReTEM}}
\newcommand{\ease}{\textsc{EASE}}

\title{EgoSIS: From Factorized Visual Ego-Transitions to Motion-Canonical Spatial Evidence for UAV Reasoning}

\name{\begin{tabular}{@{}c@{}}
Jingpu Yang$^{1,2,\ast}$, Fengxian Ji$^{2,\ast}$,
Mingxuan Cui$^{3}$, Yilin Sun$^{1}$,\\
Hang Zhang$^{4}$, Jianhua Zhu$^{1}$, and Yufeng Wang$^{1,\dagger}$
\end{tabular}}
\address{
$^{1}$Beihang University, Beijing, China\\
$^{2}$Zhongguancun Academy, Beijing, China\\
$^{3}$Northeastern University, Shenyang, China\\
$^{4}$Technology and Engineering Center for Space Utilization, Chinese Academy of Sciences,\\
Beijing 100094, China\\
$^{\ast}$Equal contribution. $^{\dagger}$Corresponding author: Yufeng Wang (\textnormal{wyfeng@buaa.edu.cn}).}

\begin{document}
\ninept  
\maketitle

\begin{abstract}
UAV video question answering requires separating camera motion from changes in
the scene, but RGB-only multimodal models receive no explicit, stable reference
for that separation.  We present EgoSIS, a pose-free adapter that converts
RGB-derived bidirectional flow into motion-canonical visual evidence in three
stages.  Factorized Visual Ego-Transitions (FVET) fits a robust image-plane
transition and exposes motion, residual-support, and reliability factors.
Reliability-Gated Ego-Transition Memory (ReTEM) uses reliability-weighted
updates for a bounded history and re-anchors it at cuts or sustained uncertainty.  Ego-Aligned
Spatial Evidence (EASE) warps supported visual features into each segment's
local anchor and injects four spatial evidence tokens per visual slice through
zero-initialized
residuals, without changing Qwen's visual-token count.  On SIS-Bench,
EgoSIS-8B obtains 89.9\% perception, 82.5\% perception-plus-memory, and
76.2\% overall accuracy, with the largest gains concentrated in
self-awareness perception and memory.  The adapter thus provides an
interpretable interface between optical flow and spatial reasoning.
\end{abstract}


\section{Introduction}
\label{sec:intro}

\begin{table*}[t]
  \centering
  \caption{Performance on SIS-Bench. Accuracy (\%) across 13 tasks for
  proprietary, open-source, and EgoSIS models.}
  \label{tab:main}
  \resizebox{\textwidth}{!}{%
  \begin{tabular}{lcccccccccccccccc}
    \toprule
    Model & Perc. & Perc.+Mem. & Overall
      & \multicolumn{8}{c}{Spatial Cognition}
      & \multicolumn{5}{c}{Self-Awareness} \\
    \cmidrule(lr){5-12}\cmidrule(lr){13-17}
      & & & & OE & OA & RD & LO & LR & PR & SC & STC
      & AR & AS & ARec & AP & PP \\
    \midrule

    \multicolumn{17}{c}{\textit{Proprietary models}} \\
    \midrule
    Gemini-3-Flash
      & 79.1 & 74.4 & 71.6 & 97.2 & 80.6 & 75.0 & 94.4 & 84.0 & 89.7 & 71.8 & \textbf{57.7} & 66.5 & 84.1 & 42.4 & 61.6 & 53.7 \\
    Kimi-2.5
      & 73.0 & 73.2 & 71.0 & 97.0 & 78.0 & \textbf{80.5} & \textbf{98.0} & 81.9 & 87.3 & \textbf{76.4} & 53.1 & 50.9 & 77.8 & 53.0 & 65.4 & 57.0 \\
    Doubao-Seed-1.8
      & 68.7 & 73.3 & 70.6 & 97.4 & 72.6 & 76.5 & 96.1 & \textbf{84.4} & 87.7 & 66.2 & 57.3 & 43.7 & 85.1 & \textbf{59.2} & 63.1 & 54.0 \\
    Qwen3.5-Plus
      & 73.3 & 71.7 & 70.1 & 97.6 & 76.2 & 78.5 & 97.7 & 83.3 & 87.7 & 75.9 & 54.8 & 52.8 & 82.5 & 42.4 & \textbf{69.2} & 58.5 \\
    GPT-5.4
      & 73.7 & 72.2 & 70.0 & \textbf{98.4} & 74.9 & 67.0 & 94.8 & 78.6 & 86.9 & 64.1 & 57.3 & 57.3 & 77.8 & 49.9 & 65.8 & \textbf{58.8} \\

    \midrule
    \multicolumn{17}{c}{\textit{Open-source baselines}} \\
    \midrule
    Qwen3-VL-8B-Instruct
      & 74.8 & 67.0 & 63.1 & 97.2 & 82.2 & 73.5 & 95.8 & 74.9 & 82.9 & 54.4 & 51.0 & 55.1 & 60.3 & 32.2 & 57.0 & 29.0 \\
    Qwen3-VL-8B-Thinking
      & 72.2 & 66.5 & 64.6 & 95.5 & 76.0 & 72.5 & 95.1 & 78.6 & 79.8 & 64.6 & 55.6 & 53.4 & 62.9 & 33.6 & 64.6 & 46.3 \\
    Qwen3-VL-4B-Instruct
      & 73.3 & 65.0 & 62.5 & 96.3 & 80.4 & 71.0 & 93.8 & 70.7 & 82.1 & 73.3 & 47.3 & 53.4 & 61.3 & 29.1 & 57.0 & 36.4 \\
    InternVL3.5-8B
      & 73.4 & 64.0 & 61.1 & 96.1 & 76.5 & 70.5 & 86.3 & 72.0 & 73.8 & 55.4 & 47.3 & 56.1 & 52.4 & 31.8 & 55.1 & 41.9 \\
    Qwen3-VL-30B-A3B-Instruct
      & 68.7 & 62.4 & 61.0 & 97.2 & 78.6 & 79.0 & 95.1 & 76.5 & 85.3 & 67.7 & 47.7 & 39.7 & 44.8 & 28.4 & 59.7 & 48.9 \\
    GLM-4.1V-9B-Thinking
      & 69.5 & 62.9 & 60.4 & 92.1 & 77.5 & 72.0 & 93.5 & 64.1 & 75.4 & 63.6 & 49.8 & 48.1 & 55.9 & 34.8 & 48.7 & 43.4 \\
    MiMo-VL-7B-RL
      & 68.8 & 61.1 & 59.2 & 96.3 & 74.2 & 73.5 & 80.1 & 75.2 & 81.3 & 72.3 & 47.3 & 44.8 & 43.2 & 29.7 & 51.7 & 40.8 \\
    Qwen2.5-VL-7B-Instruct
      & 67.1 & 57.5 & 55.8 & 96.5 & 71.1 & 69.0 & 73.2 & 54.2 & 69.0 & 67.7 & 41.5 & 43.3 & 55.6 & 29.2 & 60.8 & 31.6 \\
    Qwen2.5-VL-3B-Instruct
      & 58.6 & 56.2 & 53.6 & 92.7 & 48.8 & 64.5 & 90.8 & 51.9 & 62.7 & 57.4 & 42.7 & 37.9 & 62.5 & 35.6 & 53.6 & 23.2 \\
    \midrule
    \multicolumn{17}{c}{\textit{EgoSIS (ours)}} \\
    \midrule
    EgoSIS-3B
      & 84.1 & 74.9 & 68.2 & 96.3 & 69.5 & 66.5 & 92.8 & 68.8 & 72.6 & 58.5 & 44.0 & 88.6 & 81.6 & 49.3 & 46.8 & 22.4 \\
    EgoSIS-4B
      & 87.3 & 79.8 & 73.0 & 97.0 & 80.9 & 76.0 & 95.4 & 79.5 & 82.5 & 60.5 & 43.6 & 87.2 & 83.5 & 55.2 & 57.0 & 27.2 \\
    EgoSIS-7B
      & 87.6 & 79.8 & 73.1 & 97.4 & 79.1 & 73.5 & 94.1 & 79.2 & 75.8 & 59.0 & 46.1 & \textbf{89.7} & 84.1 & 57.1 & 57.4 & 26.1 \\
    EgoSIS-8B
      & \textbf{89.9} & \textbf{82.5} & \textbf{76.2} & 97.8 & \textbf{86.6} & 78.0 & \textbf{98.0} & 79.9 & \textbf{90.5} & 64.6 & 53.1 & \textbf{89.7} & \textbf{87.3} & 57.5 & 58.6 & 31.2 \\

    \bottomrule
  \end{tabular}}
\end{table*}

UAV autonomy spans resilient communication, including agent-based
anti-jamming~\cite{yang2025antijamming}, LLM-assisted frequency-game
planning~\cite{yang2025uavfpg}, and risk-sensitive anti-spoofing~\cite{zhang2026idtd},
as well as visual navigation~\cite{wang2024traveluav}.
Here we study UAV video reasoning, which must model both the scene and the
motion of the observing platform.  Camera motion shifts most image points, so apparent displacement
entangles ego-motion, independently moving objects, occlusion, and artifacts.
An RGB-only video multimodal large language model (MLLM)~\cite{videochatgpt,video_llava,videollama,timechat,videomme,spatialvlm} has no explicit stable
reference for this ambiguity, which is especially harmful for temporal spatial
questions.
Recent UAV video benchmarks such as SIS-Bench~\cite{sisbench} quantify this gap:
the closed-source models in Table~\ref{tab:main} remain below 75\% overall
accuracy, with limited performance on spatio-temporal consistency and action
recall under sustained platform motion.

Related visual work uses geometric gating for temporally stable UAV
segmentation~\cite{yang2026geometricgating} and reliable local correspondence
for multimodal UAV fusion in GAAT~\cite{yang2026gaat,superglue}.
GeoCoT~\cite{song2025geocot} combines contextual and spatial clues for image
geolocation; our focus is a stable reference across moving UAV views.
Optical flow provides cues for action and camera-motion recognition~\cite{sisbench,raft,pwcnet,flownet2,liteflownet,gma},
but a dense, view-dependent flow
map does not by itself separate global from residual motion, identify unreliable
correspondences, or define comparable coordinates across segments.
We therefore adopt one causal chain: factorize each transition, retain reliable
history, and align supported evidence to a local anchor.

\begin{figure}[t]
  \centering
  \includegraphics[width=0.95\columnwidth, trim=0 0.5cm 0 0.5cm, clip]{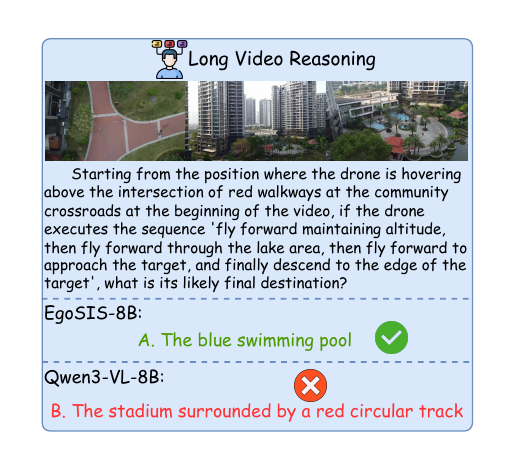}
  \caption{Qualitative comparison on a long UAV video: \method{}-8B identifies
  the destination swimming pool, whereas Qwen3-VL-8B predicts a stadium.}
  \label{fig:qualitative_reasoning}
\end{figure}

In Fig.~\ref{fig:qualitative_reasoning}, \method{}-8B follows the trajectory
to a swimming pool, whereas Qwen3-VL-8B predicts a stadium, motivating a
motion-canonical reference for temporal spatial reasoning.

\method{} is a pose-free RGB adapter that implements this chain.  Factorized
Visual Ego-Transitions (\fvet) checks bidirectional flow and emits motion,
residual/static-support, and reliability factors.  Reliability-Gated
Ego-Transition Memory (\retem) keeps a bounded history and composes only safe
geometry.  Ego-Aligned Spatial Evidence (\ease) warps supported features into
the current segment anchor.  The representation is an image-plane proxy, not
metric pose or a 3-D map; zero-initialized residuals preserve Qwen's visual
token count, and F/FR/FRE denote the cumulative variants.

Our contributions are threefold:
\begin{itemize}
  \item We introduce \fvet, which factorizes bidirectional visual transitions
  into global image-plane motion, residual/static support, and reliability.
  \item We propose \retem, a bounded transition memory with reliability gates,
  cut-aware segmentation, and safe geometric re-anchoring.
  \item We develop \ease, which converts reliable transition history into local
  motion-canonical evidence without expanding the visual-token sequence, and
  evaluate the complete adapter on SIS-Bench.
\end{itemize}

\section{EgoSIS}
\label{sec:method}

\subsection{Overview}
\label{ssec:overview}

\begin{figure*}[t]
  \centering
  \includegraphics[width=\textwidth, trim=0 0 0 14cm, clip]{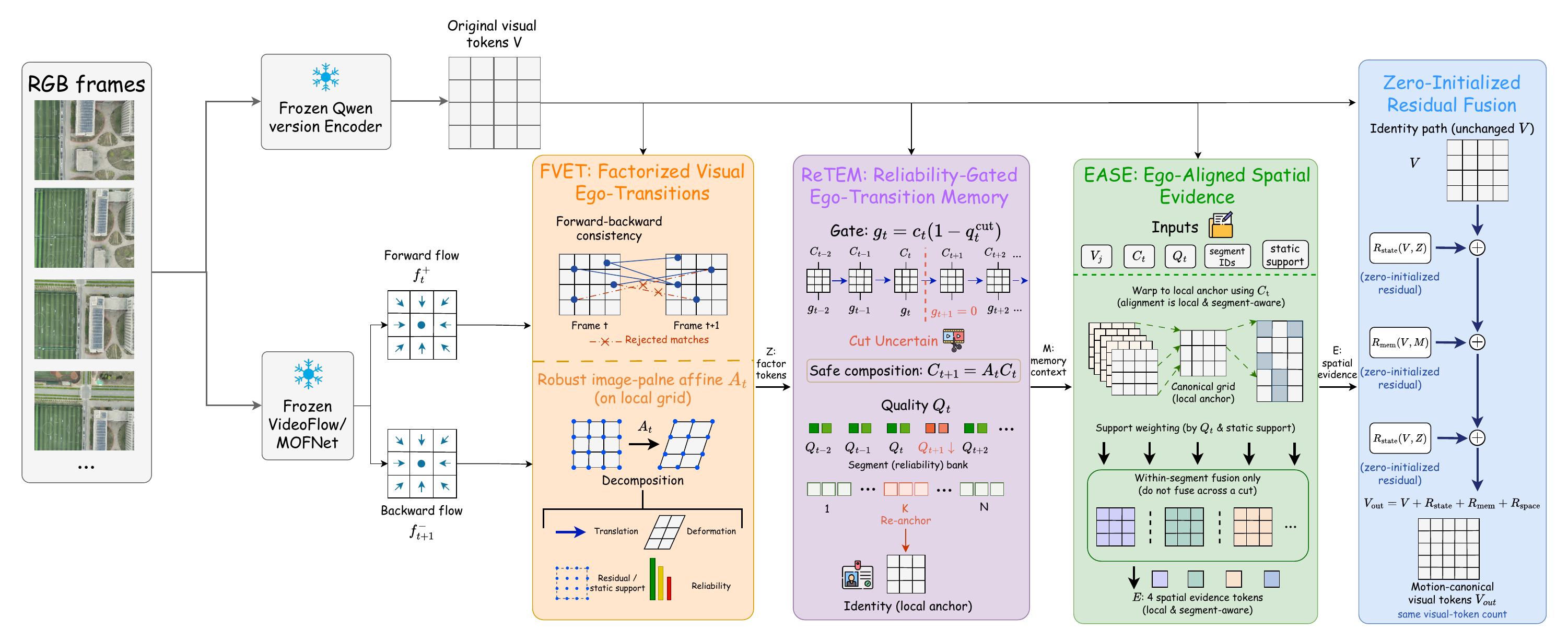}
  \caption{Overview of \method{}, a pose-free RGB adapter for UAV video
  reasoning.  \fvet{} factorizes bidirectional flow, \retem{} gates and
  re-anchors transition memory, and \ease{} aligns visual evidence to segment
  anchors.  The resulting contexts $Z$, $M$, and $E$ update frozen Qwen visual
  features through zero-initialized residuals without changing the visual-token count.}
  \label{fig:overview}
\end{figure*}

Let $\{I_t\}_{t=0}^{T-1}$ denote $T$ RGB frames sampled from a UAV video and
let $V\in\mathbb{R}^{G_t\times G_h\times G_w\times d}$ be the visual-token
grid produced by the frozen Qwen vision encoder~\cite{qwen25vl,qwen3vl} after its spatial merger,
where $G_t,G_h,G_w$ are its temporal and spatial grid sizes and $d$ is its
feature width.  A frozen VideoFlow/MOFNet estimator~\cite{shi2023videoflow} produces bidirectional
flow.  As Fig.~\ref{fig:overview} shows, \fvet{} factorizes each flow pair,
\retem{} gates packet-history updates and composes safe geometry, and
\ease{} aligns supported visual features to segment anchors.  We
denote \fvet{}, \fvet{}+\retem{}, and the full model by F, FR, and FRE,
respectively.  Boundary duplication aligns flow entry $t$ with source frame
$t$; the final outgoing edge is masked as invalid.

The two frozen streams yield visual tokens $V$ and flow pairs.
\fvet{} extracts factor tokens $Z$; \retem{} maintains reliable segment
history $M$; \ease{} pools four aligned evidence tokens $E$.
These contexts update $V$ through zero-initialized residuals.

\subsection{Factorized Visual Ego-Transitions}
\label{ssec:fvet}

For edge $t$, let $f_t^+$ map frame $t$ to $t+1$ and let $f_{t+1}^-$ be the
backward flow of the latter frame.  For normalized grid coordinate
$p\in[-1,1]^2$, let $S=\operatorname{diag}(2/(W-1),2/(H-1))$ convert pixel flow
on an $H\times W$ grid to normalized displacement.  The forward endpoint and
forward--backward consistency error~\cite{meister2018unflow} are
\begin{gather}
\hat p_t(p)=p+S f_t^+(p),
\label{eq:fb_endpoint}\\
e_t(p)=
\left\|f_t^+(p)
+\mathcal{W}\!\left(f_{t+1}^-;\hat p_t(p)\right)\right\|_2 .
\label{eq:fb_error}
\end{gather}
where $\mathcal{W}(b;q)$ bilinearly samples $b$ at $q$.  Only $\hat p_t$ is
normalized; $f_t^+$, $e_t$, and the threshold remain in pixels.  We retain
finite correspondences with in-bounds endpoints, finite backward support, and
$e_t(p)\leq\tau_t(p)$, where
$\tau_t(p)=0.5+0.05(\|f_t^+(p)\|_2+
\|\mathcal{W}(f_{t+1}^-;\hat p_t(p))\|_2)$.

We fit an affine $A_t\in\mathbb{R}^{3\times3}$ from frame $t$ to $t+1$
in normalized coordinates using Huber IRLS~\cite{huber1964robust} with MAD
rejection.  If too few consistent
points survive, the edge remains time-aligned but contributes zero support and
confidence.  The resulting packet is
\begin{equation}
\resizebox{0.90\columnwidth}{!}{$
s_t=[\Delta x_t/W,\Delta y_t/H,\theta_t,\log\alpha_t,r_t^{\rm med},
r_t^{\rm MAD},c_t,q_t^{\rm cut},\Delta t_t]
$}
\label{eq:motion_packet}
\end{equation}
where $\Delta x_t,\Delta y_t$ are
pixel-equivalent affine translations, $\theta_t$ and $\alpha_t$ are rotation
and geometric-mean absolute scale, and $r_t^{\rm med},r_t^{\rm MAD}$ are
residual statistics normalized by the flow-grid diagonal; $\Delta t_t$ is the
timestamp gap (or frame-index gap when timestamps are unavailable).  Confidence $c_t$
combines valid support, forward--backward agreement, residual/static support,
and affine conditioning; cut score $q_t^{\rm cut}$ also uses RGB photometric
warp disagreement.  FVET projects the translation, deformation,
residual/support, and reliability factors of $s_t$ in Eq.~\eqref{eq:motion_packet}
with relative-time and factor-type embeddings, yielding motion tokens $Z=\{Z_t\}$.

\subsection{Reliability-Gated Ego-Transition Memory}
\label{ssec:retem}

Let $h_t$ be the semantic state before edge $t$, $\psi$ its learned encoder,
and $\mathbf{1}[\cdot]$ an indicator.  ReTEM reliability-weights the update as
\begin{gather}
g_t = c_t(1-q_t^{\mathrm{cut}})
\,\mathbf{1}[\text{edge }t\text{ valid}],
\label{eq:retem_gate}\\
h_{t+1} = (1-g_t)h_t
+ g_t\,\operatorname{GRU}\!\left(\psi(s_t),h_t\right).
\label{eq:retem_update}
\end{gather}
Within a segment, $C_t\in\mathbb{R}^{3\times3}$ maps the current anchor to
frame $t$, while $Q_t\in[0,1]$ records cumulative geometric quality; both start
at $(I,1)$.  With geometry threshold $\gamma_{\rm geo}$, a safe candidate is
composed as
\begin{equation}
 (C_{t+1},Q_{t+1})=
 \begin{cases}
 (A_tC_t,Q_tg_t), & g_t\geq\gamma_{\rm geo},\ A_tC_t\ {\rm safe},\\
 (C_t,Q_t), & \text{otherwise.}
 \end{cases}
 \label{eq:retem_geometry}
\end{equation}
The order $A_tC_t$ follows the anchor-to-frame direction.  A hard cut,
persistent low confidence, an unsafe candidate, a post-update quality drop, or
the segment horizon re-anchors the track at $(I,1)$.  Sub-threshold cut evidence
only soft-gates the semantic state; a threshold crossing starts a new segment.
The final sentinel is skipped, and memory becomes valid only after a
positive-confidence valid edge.  Mean-pooled past segments form a bounded bank;
learned queries attend to it and the current history to produce context $M$.

\subsection{Ego-Aligned Spatial Evidence}
\label{ssec:ease}

For contiguous Qwen temporal patch $j$, let
$V_j\in\mathbb{R}^{G_h\times G_w\times d}$ be its visual map, $r_j$ its first
frame, and $s_j$ its ReTEM segment.  Since
$C_{r_j}$ maps anchor to frame, $\operatorname{warp}(X,C_{r_j})$ samples
current-grid tensor $X$ into the anchor.  For conservative static mask $m_j$,
\begin{equation}
 w_j=\operatorname{warp}(m_j,C_{r_j})\odot u_j\odot v_j^{\rm stat}\odot
 c_j^{\min}(1-q_j^{\max})Q_{r_j},
 \label{eq:ease_compact}
\end{equation}
Here $u_j$ and $v_j^{\rm stat}$ are in-bounds feature and warped-static
support; $c_j^{\min}$ and $q_j^{\max}$ aggregate valid internal edges whose
endpoints lie in the patch, and $Q_{r_j}$ is cumulative quality.  The mask
$m_j$ intersects their static support.
Groups without internal edges use $c_j^{\min}=1,q_j^{\max}=0$; those crossing
a cut, segment boundary, or pure padding are invalid.  Valid groups in segment
$s$ form
\begin{equation}
 \bar V_s=\frac{\sum_{j\in s}w_j\odot
 \operatorname{warp}(V_j,C_{r_j})}{\sum_{j\in s}w_j+\epsilon};
 \label{eq:ease_fusion_compact}
\end{equation}
where $\epsilon$ and a masked fallback keep empty support finite.  A
$2\times2$ pool yields four internal context tokens $E$; contexts $Z$, $M$,
and $E$ attend to each visual slice through zero-initialized residuals:
\begin{equation}
\resizebox{0.90\columnwidth}{!}{$
V^{\mathrm{out}}
=V+R_{\mathrm{state}}(V,Z)
+R_{\mathrm{mem}}(V,M)
+R_{\mathrm{space}}(V,E)
$}
\label{eq:residual_compact}
\end{equation}
Each $R$ uses a zero-initialized output projection; validity masks suppress
unsupported fallbacks, so $V^{\mathrm{out}}$ preserves $V$'s shape and token
positions.

\section{Experiments}
\label{sec:experiments}

\subsection{Experimental Setup}
\label{ssec:setup}

Unless otherwise noted, we train each structural variant for one epoch on
SIS-Motion-54K~\cite{sisbench} using causal language-modeling loss on assistant answer tokens.
The pretrained Qwen backbone and the MOFNet-based flow estimator remain frozen;
only the EgoSIS connector and LoRA adapters~\cite{hu2022lora} on the language self-attention
Q/K/V/O projections are trained.  LoRA uses rank $32$, alpha $64$, and dropout
$0.05$.  The formal stagewise run uses a frozen census of 54,298 valid
 training instances from 11,763 videos, global batch size $32$, BF16, and
 ZeRO-2.  Videos are sampled at 2 FPS and clamped to 8--32 frames; frame
 manifests, prompts, decoding, and score denominators are held fixed within
 each comparison.

\subsection{Main Results}
\label{ssec:main_results}

We report all 13 official SIS-Bench tasks.  UAV benchmarks such as VisDrone~\cite{zhu2018visdrone}, UAVid~\cite{uavid}, and StateBench~\cite{gaat} emphasize the viewpoint and motion conditions relevant here. Under spatial cognition, OE, OA,
RD, LO, LR, PR, SC, and STC denote Object Existence, Object Attribute,
Relative Direction, Landmark Order, Landmark Recall, Positional Relationship,
Spatial Consistency, and Spatio-Temporal Consistency.  Under self-awareness,
AR, AS, ARec, AP, and PP denote Action Recognition, Action Sequence, Action
Recall, Action Prediction, and Path Planning.

Following the SIS-Bench evaluation protocol~\cite{sisbench}, Perc. covers
the perception tasks OE, OA, RD, and AR, while Perc.+Mem. additionally
includes all memory tasks.  Overall covers all 4,856 questions across
perception, memory, and reasoning.  Each aggregate accuracy is the total
number of correct answers divided by the number of questions in that group.

\begin{figure}[t]
  \centering
  \includegraphics[width=\columnwidth]{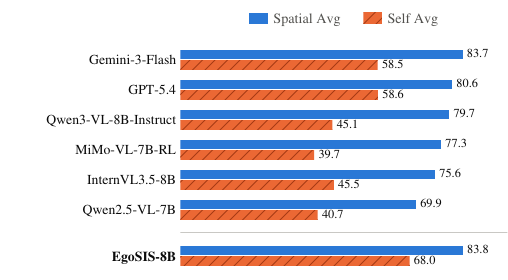}
  \caption{Question-weighted dimension accuracy using counts reconstructed
  from Table~\ref{tab:main}.}
  \label{fig:dimavg}
\end{figure}

EgoSIS-8B achieves the highest overall accuracy in Table~\ref{tab:main},
reaching 76.2\% compared with 63.1\% for Qwen3-VL-8B-Instruct.
Its largest improvements over this backbone are in recognizing the UAV's
actions and recalling their temporal history, spanning both perception and
memory within self-awareness.  The gains in reasoning are less consistent:
spatial consistency improves substantially, whereas spatio-temporal
consistency and action prediction improve only modestly.  Path planning
remains the weakest task, suggesting that better recognition and recall of
past motion do not yet translate into comparable gains in planning.

\subsection{Cross-Benchmark Generalization}
\label{ssec:generalization}

Table~\ref{tab:external} reports transfer beyond SIS-Bench.  OpenUAV-QA~\cite{sisbench},
derived from OpenUAV/TravelUAV~\cite{wang2024traveluav}, uses S/L splits at
45 frames before the model's frame cap.  CameraBench~\cite{lin2025camerabench} reports
classification mAP and VQA accuracy; MotionBench~\cite{hong2025motionbench} reports camera-motion (CM)
and overall accuracy at 32 frames.  EgoSIS-8B leads the reported OpenUAV-QA
columns at 97.3\%/93.8\%/95.6\% and CameraBench classification at 44.6\%.
It also leads MotionBench CM at 59.0\% and ties Ovis2.5-9B at 29.7\% Overall
at the displayed precision.  CameraBench VQA remains weaker: 53.6\% versus
Ovis2.5-9B's 61.5\%.  Comparisons with EgoSIS-7B involve different backbone
generations and do not isolate model scale.

\begin{table}[t]
  \centering
  \caption{Cross-benchmark generalization (\%). OU, CB, and MB denote
  OpenUAV-QA, CameraBench, and MotionBench@32f; column maxima are bold.}
  \label{tab:external}
  \setlength{\tabcolsep}{2pt}
  \renewcommand{\arraystretch}{0.82}
  \resizebox{\columnwidth}{!}{%
  \begin{tabular}{@{}l*{7}{c}@{}}
    \toprule
    & \multicolumn{3}{c}{OU} & \multicolumn{2}{c}{CB}
    & \multicolumn{2}{c}{MB} \\
    \cmidrule(lr){2-4}\cmidrule(lr){5-6}\cmidrule(lr){7-8}
    Model & S & L & All & Cl. & VQA & CM & All \\
    \midrule
    Qwen2.5-VL-7B-Instruct & 75.6 & 70.6 & 73.1 & 36.0 & 57.4 & 49.1 & 27.6 \\
    Qwen3-VL-8B-Instruct   & 82.6 & 78.5 & 80.6 & 42.3 & 55.8 & 44.9 & 24.3 \\
    InternVL3.5-8B         & 82.5 & 82.8 & 82.7 & 33.2 & 53.5 & 48.3 & 27.2 \\
    InternVL3.5-4B         & 78.3 & 73.3 & 75.8 & 31.6 & 48.4 & 52.2 & 27.3 \\
    GLM-4.1V-9B-Thinking   & 85.3 & 78.7 & 82.0 & 36.4 & 51.5 & 52.5 & 28.8 \\
    MiMo-VL-7B-RL          & 70.3 & 73.4 & 71.8 & 25.2 & 50.6 & 54.0 & 29.5 \\
    GLM-4.6V-Flash-9B      & 88.5 & 82.6 & 85.6 & 41.4 & 50.5 & 49.9 & 29.4 \\
    Kimi-VL-A3B-Instruct   & 86.3 & 85.3 & 85.8 & 34.1 & 56.3 & 46.8 & 28.6 \\
    Ovis2.5-9B             & 77.2 & 79.9 & 78.5 & 33.9 & \textbf{61.5} & 54.5 & \textbf{29.7} \\
    Qwen3-VL-8B-Thinking   & 82.9 & 78.7 & 80.8 & 42.5 & 54.6 & 56.6 & 29.3 \\
    VST-7B-RL              & 48.5 & 52.2 & 50.3 & 35.9 & 61.0 & 56.9 & 28.5 \\
    Step3-VL-10B           & 36.2 & 37.7 & 37.0 & 19.4 & 50.0 & 26.5 & 20.1 \\
    InternVL3-9B           & 86.0 & 87.6 & 86.8 & 34.4 & 46.1 & 46.2 & 27.9 \\
    MOFNet-7B              & 97.2 & 91.9 & 94.6 & 43.9 & 56.6 & 56.1 & 28.4 \\
    \midrule
    \method-7B             & 97.1 & 93.0 & 95.1 & 36.6 & 52.8 & 57.1 & 28.4 \\
    \method-8B             & \textbf{97.3} & \textbf{93.8} & \textbf{95.6} & \textbf{44.6} & 53.6 & \textbf{59.0} & \textbf{29.7}\\
    \bottomrule
  \end{tabular}}
\end{table}

\subsection{Ablation and Analysis}
\label{ssec:ablation}

Table~\ref{tab:ablation} compares 8B variants on 4,856 SIS-Bench questions.
Qwen3-VL-8B-Instruct is the zero-shot reference.  Visual-only SFT is compared
with F, FR, and FRE, which successively introduce FVET, ReTEM, and EASE.
FR freezes the inherited
F model and trains ReTEM; FRE freezes FR and trains the zero-initialized
EASE branch.  These comparisons measure incremental stagewise training
effects, with additional optimization at each stage.

\begin{table}[t]
  \centering
  \caption{SIS-Bench component ablation (\%) with matched evaluation:
  4,856 questions overall and task-specific denominators.}
  \label{tab:ablation}
  \resizebox{\columnwidth}{!}{%
  \begin{tabular}{lcccccccc}
    \toprule
    Variant & \multicolumn{2}{c}{Spatial Reasoning}
      & \multicolumn{5}{c}{Self-Awareness} & Overall \\
    \cmidrule(lr){2-3}\cmidrule(lr){4-8}
      & SC & STC & AR & AS & ARec & AP & PP & \\
    \midrule
    Zero-shot reference
      & 54.4 & 51.0 & 55.1 & 60.3 & 32.2 & 57.0 & 29.0 & 63.1 \\
    Visual-only SFT
      & 56.4 & 50.6 & 56.0 & 87.0 & 47.8 & 55.9 & 29.0 & 65.2 \\
    EgoSIS-F
      & 62.1 & 51.0 & 77.3 & 85.1 & 53.2 & 54.4 & 29.8 & 73.3 \\
    EgoSIS-FR
      & \textbf{65.1} & 52.3 & 89.4 & 86.3 & 52.6 & 53.6 & 30.1 & 74.1 \\
    EgoSIS-FRE
      & 64.6 & \textbf{53.1} & \textbf{89.7} & \textbf{87.3} & \textbf{57.5} & \textbf{58.6} & \textbf{31.2} & \textbf{76.2} \\
    \bottomrule
  \end{tabular}}
\end{table}

Visual-only SFT mainly improves action sequence understanding and recall,
with limited gains in action recognition and spatial reasoning.  Introducing
FVET substantially improves action recognition and spatial consistency,
and ReTEM further strengthens both.  These gains do not extend uniformly
to other tasks: action recall and prediction decline slightly when ReTEM
is added.  EASE then improves recall and prediction beyond both F and FR,
consistent with the benefit of aligning visual evidence across frames.
The complete FRE model reaches 76.2\% overall accuracy, compared with
65.2\% for visual-only SFT, and performs best on six of the seven displayed
tasks.  Spatial consistency remains slightly higher with FR, indicating
a small tradeoff when aligned evidence is introduced.

\section{Conclusion}
\label{sec:conclusion}

UAV video reasoning needs a stable spatial reference because image
displacement otherwise conflates platform motion with scene change.
EgoSIS addresses this problem as a causal chain: FVET factorizes each
bidirectional transition, ReTEM gates history by reliability and re-anchors
unsafe segments, and EASE aligns supported visual evidence without changing
the number or positions of visual tokens. On SIS-Bench, EgoSIS-8B reaches
89.9\% perception, 82.5\% perception-plus-memory, and 76.2\% overall accuracy.
Cross-benchmark results for the EASE-off 8B setting show a clear gain on
OpenUAV-QA but mixed performance on CameraBench and MotionBench. The current
representation remains an image-plane proxy rather than metric pose, and
checkpoint-controlled progressive ablations are still needed to isolate the
contributions of ReTEM and EASE.

\bibliographystyle{IEEEbib}
\bibliography{refs}

\end{document}